%% file: colm2026_conference.tex
\documentclass{article} 
\usepackage[preprint]{colm2026_conference}

\usepackage{microtype}
\usepackage{hyperref}
\usepackage{url}
\usepackage{booktabs}
\usepackage{amsmath}
\usepackage[normalem]{ulem}

\usepackage{multirow}
\usepackage[table]{xcolor}

\usepackage[most]{tcolorbox}
\usepackage{xcolor}
\definecolor{promptblue}{HTML}{34495E}
\definecolor{promptgray}{HTML}{F3F3F3}

\usepackage{wrapfig}

\usepackage{amssymb} 
\usepackage{booktabs}

\usepackage{float}

\usepackage{lineno}

\definecolor{darkblue}{rgb}{0, 0, 0.5}
\hypersetup{colorlinks=true, citecolor=darkblue, linkcolor=darkblue, urlcolor=darkblue}

\title{ContextBudget: Budget-Aware Context Management for Long-Horizon Search Agents}

\author{
  Yong Wu$^{1}$\thanks{Equal contribution}\quad
  YanZhao Zheng$^{2}$\footnotemark[1]\quad
  TianZe Xu$^{2}$\quad
  \AND
  ZhenTao Zhang$^{2}$\quad
  YuanQiang Yu$^{2}$\quad
  JiHuai Zhu$^{2}$\quad
  Chao Ma$^{2}$ 
  \AND
  BinBin Lin$^{1}$\quad
  BaoHua Dong$^{2}$\quad
  HangCheng Zhu$^{2}$\quad
  RuoHui Huang$^{2}$\quad
  Gang Yu$^{2}$ \\[4pt]
  $^{1}$Zhejiang University, Hangzhou, China \quad 
  $^{2}$Alibaba Group, Hangzhou, China \\
  {\small\texttt{wu.yong@zju.edu.cn, binbinlin@zju.edu.cn}}\\
  {\small\texttt{\{zhengyanzhao.zyz, xutianze.xtz, zhangzhentao.zzt, yuyuanqiang.yyq,}}\\
  {\small\texttt{zhujihuai.zjh, mc524716, baohua.dbh, linran.lr09, wentong,}}\\
  {\small\texttt{ruohui.huang, gang.yu\}@alibaba-inc.com}}
}

\usepackage{booktabs}      
\usepackage{multirow}      
\usepackage{graphicx}      
\usepackage{xcolor}        
\usepackage{array}         
\usepackage{makecell}

\usepackage{fancyhdr}
\usepackage{graphicx}

\fancypagestyle{logoheader}{
    \fancyhead[R]{\includegraphics[height=0.6cm, width=3cm, keepaspectratio]{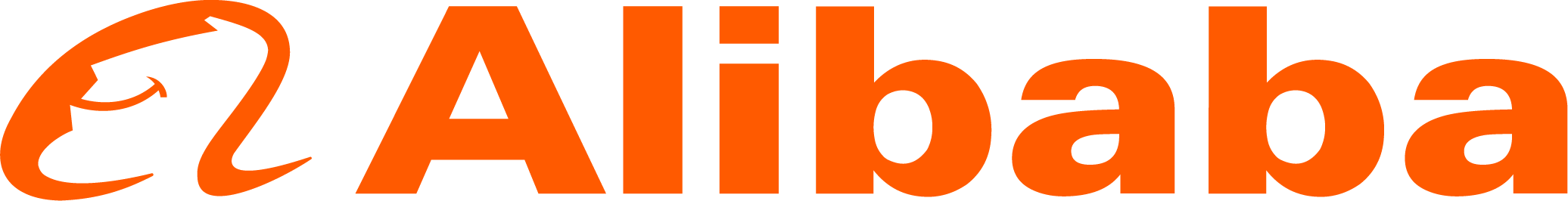}}
}

\begin{document}

\ifcolmsubmission
\linenumbers
\fi

\maketitle






\begin{abstract}
LLM-based agents show strong potential for long-horizon reasoning, yet their context size is limited by deployment factors (e.g., memory, latency, and cost), yielding a constrained context budget. As interaction histories grow, this induces a trade-off between retaining past information and staying within the context limit. To address this challenge, we propose Budget-Aware Context Management (BACM), which formulates context management as a sequential decision problem with a context budget constraint. It enables agents to assess the available budget before incorporating new observations and decide when and how much of the interaction history to compress. We further develop BACM-RL, an end-to-end curriculum-based reinforcement learning approach that learns compression strategies under varying context budgets. Experiments on compositional multi-objective QA and long-horizon web browsing benchmarks show that BACM-RL consistently outperforms prior methods across model scales and task complexities, achieving over $1.6\times$ gains over strong baselines in high-complexity settings, while maintaining strong advantages as budgets shrink, where most methods exhibit a downward performance trend. 
\end{abstract}

\input{section/Introduction}

\input{section/RelatedWork}

\input{section/Methodology}

\input{section/Experiment}

\input{section/Conclusions}

\bibliography{colm2026_conference}
\bibliographystyle{colm2026_conference}

\appendix
\input{section/Appendix}

\end{document}

%% file: section/Introduction.tex
\section{Introduction}
\label{sec:introduction}
The growing capabilities of LLM agents and their increasing adoption in long-horizon interactive applications have made effective context management a pressing requirement\citep{hua2025contextengineering20context,luo2025largelanguagemodelagent}.
As agents increasingly operate over extended trajectories, interaction histories rapidly accumulate and cause substantial context expansion\citep{yao2023reactsynergizingreasoningacting}.
At the same time, the maximum context window remains strictly bounded by deployment resources such as memory footprint, inference latency, and serving cost\citep{liu2024lost,packer2024memgptllmsoperatingsystems}.
Consequently, controlling context growth has emerged as a central concern for sustaining reliable long-horizon reasoning.

Recent agent systems therefore increasingly adopt context compression\citep{zhou2025mem1learningsynergizememory,wu2025resumunlockinglonghorizonsearch, hua2025contextengineering20context}, where past observations, intermediate traces, and retrieved evidence are summarized or rewritten during inference.
This paradigm has gained prominence due to its empirical effectiveness and conceptual simplicity: it preserves task-critical information under fixed context limits while requiring no external memory or architectural modification\citep{hu2026memoryageaiagents}.

However, most existing compression methods adopt a budget-free formulation\citep{zhou2025mem1learningsynergizememory,wu2025resumunlockinglonghorizonsearch, hua2025contextengineering20context}, treating compression as a static operation without explicit conditioning on the available context budget.
This simplification introduces two critical failure modes.
Under relaxed budgets, agents may over-compress and erase research-critical evidence, reducing information fidelity\citep{li2023compressing,tang2025lciteeval}.
Under tight budgets, agents may under-compress and overflow the context limit, causing truncation or brittle reasoning failures\citep{liu2024lost,hsieh2024ruler}.

Recent work has begun to incorporate budget awareness into agentic reasoning, studying constraints such as tool-call budgets and output token budgets \citep{liu2025budget_tooluse, tale2024tokenbudget}.
These methods highlight that interaction scaling is limited by finite resources in deployment\citep{snell2024scaling}.
However, they primarily regulate budgets through external controls on computation or action counts.
They do not systematically address proactive history compression when the context-window budget itself is the limiting resource.

In this paper, we propose Budget-Aware Context Management, a framework enabling LLM agents to perform long-horizon reasoning under explicit context-window budgets. The central idea is to formulate compression in context management as a budget-constrained sequential decision problem, allowing compression decisions to adapt dynamically to remaining context capacity throughout the reasoning process. To achieve this, the framework introduces an explicit budget-awareness signal enabling agents to determine when to compress, how much to compress, and which information to preserve for future reasoning. This budget-conditioned control retains research-critical evidence under strict budgets while avoiding unnecessary information loss when capacity is sufficient.

To operationalize this formulation, we develop a budget-constrained reinforcement learning approach that extends Group Relative Policy Optimization (GRPO) \citep{shao2024deepseekmath} to optimize context management end to end. Instead of relying on handcrafted heuristics, the approach directly aligns compression strategies with downstream task success under explicit budget constraints. It further incorporates a progressively tightened budget curriculum and a penalty for budget violations to improve robustness under strict context limits.


Extensive experiments on compositional multi-objective QA and long-horizon web browsing benchmarks demonstrate that the approach consistently improves reasoning robustness across a wide range of context budgets, with the largest gains under stringent budget regimes. Our contributions can be summarized as follows:
\begin{itemize}
    \item We propose \textbf{Budget-Aware Context Management}, a framework that formulates budget-aware context compression for LLM agents as a sequential decision problem under explicit context-window constraints, enabling adaptive compression throughout long-horizon trajectories.
    \item We develop a budget-constrained RL method that extends GRPO with a progressively tightened budget curriculum and overflow-sensitive regularization for robust context management under strict budgets.
    \item We conduct extensive empirical studies on compositional multi-objective QA and long-horizon web browsing benchmarks, and further provide comparative analyses of compression behavior and efficiency under context constraints.

\end{itemize}

%% file: section/RelatedWork.tex
\section{Related Work}
\label{sec:related_work}


\paragraph{Context Management in LLM Agents.}
Agent reasoning accumulates prior observations and intermediate reasoning traces in the prompt over successive steps, causing linear context growth that induces the lost in the middle effect and exceeds the effective processing window \citep{yao2023reactsynergizingreasoningacting, chen2023fireactlanguageagentfinetuning, liu2024lost}. To address this, early approaches introduce external memory systems representing context as tiered storage and retrieving relevant history \citep{packer2024memgptllmsoperatingsystems, zhong2023memorybankenhancinglargelanguage, kang2025memoryosaiagent, li2025memosoperatingmemoryaugmentedgeneration}. Although these methods extend accessible context, they rely on external memory rather than updating in-context representations during reasoning. 

More recent work formulates context management as a compression problem, condensing observations and reasoning traces to maximize information density under fixed context budgets \citep{jiangLLMLinguaCompressingPrompts2023, jiangLongLLMLinguaAcceleratingEnhancing2024, chevalier2023autocompressors, mu2023gisttokens}. In single-turn settings, compression operates at the input level. In multi-turn settings, it becomes continuous state management, mapping interaction histories into bounded representations, recursively compressing trajectories into higher-level abstractions, and maintaining structured task states for stable reasoning \citep{wu2025resumunlockinglonghorizonsearch, zhou2025mem1learningsynergizememory, yu2025memagentreshapinglongcontextllm, yuan2025memsearchertrainingllmsreason, chen2025iterresearch, sun2025scalinglonghorizonllmagent, ye2025agentfold,zhang2026memoryactionautonomouscontext}. However, existing methods adopt a \textit{budget-free} formulation where compression is treated as a static or periodically triggered operation without conditioning on the available context budget. In contrast, this work models compression as a dynamic process conditioned on the context budget, enabling adaptive trade-offs between retained information and resource constraints.


\paragraph{Budget-aware Inference Scaling.}
Budget-aware inference scaling studies optimize LLM reasoning under explicit test-time resource constraints \citep{snell2024scaling}. Prior work mainly controls output length or computation, including token-budgeted reasoning and adaptive token allocation \citep{li2023compressing, tale2024tokenbudget, wen2025budgetthinkerempoweringbudgetawarellm, li2025steeringllmthinkingbudget}, showing performance depends on computation allocation but is largely limited to single-turn settings. In agent settings, recent work introduces budget-aware control over interaction processes, such as constraining tool use and tracking resource consumption in multi-step reasoning \citep{liu2025budget_tooluse}, showing performance is bounded by finite budgets.

However, these approaches primarily regulate observable interaction behaviors such as action frequency or search strategies, rather than adapting reasoning to varying context-window limits imposed by deployment constraints. Consequently, when the context window becomes the primary bottleneck, existing methods lack mechanisms to adapt context management to the available budget. In contrast, this work formulates context-window management as a budget-aware decision process, enabling agents to dynamically adjust their history representations under real-time budget signals.

%% file: section/Methodology.tex
\begin{figure*}[t]
\centering
\includegraphics[width=\textwidth]{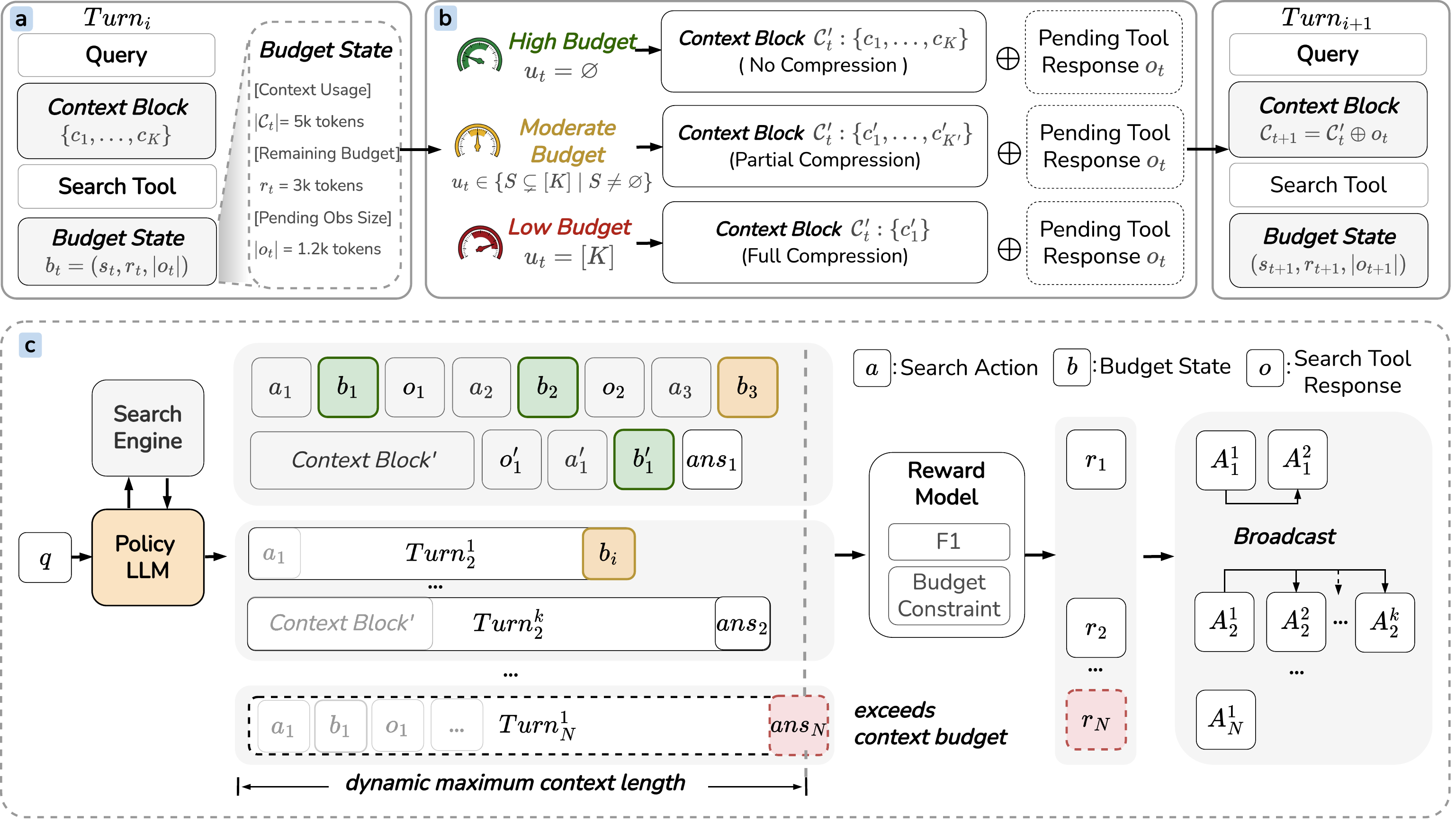}
\caption{
Overview of the proposed framework. 
(a) The agent first observes the budget-conditioned state $b_t=(s_t,r_t,|o_t|)$ before loading the pending observation. 
(b) Conditioned on $b_t$, the policy selects a refinement action $u_t$ to perform \textsc{Null}, \textsc{Partial}, or \textsc{Full} commit-block aggregation, yielding an updated context $\mathcal{C}'_t$. 
(c) The policy is trained with multi-turn GRPO under a progressively tightened budget curriculum, where only trajectories satisfying the context budget contribute reward and optimization.
}
\label{fig:pipeline}
\vspace{-10pt}
\end{figure*}

\section{Budget-Aware Context Management}
We propose a budget-aware context management framework for long-horizon search agents under strict context limits. The core idea is to formulate context management as a budget-constrained sequential decision problem. To realize this formulation, we introduce two key mechanisms: (1) a budget-conditioned inference state with deferred observation loading that exposes remaining context headroom before appending new observations (Section~\ref{sec:budget_state}), and (2) a commit-block aggregation mechanism that enables the agent to adaptively control when and how much to compress under varying budget pressures (Section~\ref{sec:commit_block}). We then optimize this decision process using a multi-turn GRPO objective with a progressively tightened context-window curriculum (Section~\ref{sec:grpo_training}). Together, these components enable effective long-horizon context allocation under strict budgets.

\subsection{Budget-Conditioned State with Deferred Loading}
\label{sec:budget_state}

We formulate context management as a budget-aware decision process to avoid both premature information loss from aggressive compression and reasoning failures due to context overflow. Instead of immediately appending new observations, the agent first evaluates available capacity and adjusts its existing context before incorporating additional information. Intuitively, the agent must reorganize its context based on how much space remains and how much space the incoming observation will require.

Formally, we extend the standard MDP with a budget-conditioned state. Let $\mathcal{S}$, $\mathcal{A}$, and $B$ denote the state space, action space, and fixed context budget. At step $t$, the agent is given an augmented state
\[
b_t = (s_t, r_t, |o_t|), \qquad r_t = B - |\mathcal{C}_t|,
\]
where $B$ denotes the total context budget, $|\mathcal{C}_t|$ denotes the token length of the current context buffer, and $r_t$ represents the remaining context budget. Here $s_t$ denotes the current reasoning state, and $|o_t|$ denotes the token length of the pending observation. The size of $o_t$ is observable, while its content remains hidden.

Given this state, the agent first determines a refinement action before accessing the new observation. Specifically, the policy samples $u_t\sim\pi_\theta(\cdot\mid b_t)$ to produce an updated context $\mathcal{C}'_t$ that satisfies the budget constraint $|\mathcal{C}'_t|\le B-|o_t|$. The observation is then appended to form the next context:
\[
\mathcal{C}_{t+1} = \mathcal{C}'_t \oplus o_t .
\]

As a result, context compression is guided by the future capacity implied by $|o_t|$, rather than a reactive post-processing step. This ordering reserves sufficient capacity for the incoming observation before incorporation, ensuring the context remains within the budget constraint.

\subsection{Commit-Block Aggregation for Budget-Aware Context Compression}
\label{sec:commit_block}

To support adaptive compression under dynamic budgets, we introduce a commit-block aggregation mechanism. It allows the agent to decide when to compress and how much to reduce based on the current budget. This avoids context overflow and unnecessary information loss. It also integrates compression into the policy instead of treating it as a separate post-processing step.

The mechanism induces three budget-dependent regimes, as illustrated in Figure~\ref{fig:pipeline}. Under high budget, the policy favors preserving the full interaction history, deferring compression to retain maximal context. Under moderate budget, it shifts toward selective aggregation, compressing redundant segments while preserving salient information. Under low budget, the policy collapses the context into a fully aggregated representation. This keeps it within budget while supporting long-horizon reasoning.

Formally, at step $t$, the context is a buffer of $K$ coherent segments, denoted by $\mathcal{C}_t=\{c_1,\dots,c_K\}$. Each segment $c_i$ is a semantically contiguous portion of the interaction history. Conditioned on the budget-aware state $b_t=(s_t,r_t,|o_t|)$, the policy samples a structured action $u_t\sim\pi_\theta(\cdot\mid b_t)$ from three mutually exclusive categories:
\[
u_t \in \underbrace{\{\emptyset\}}_{\text{\textsc{Null}}}
\;\uplus\;
\underbrace{\{\mathcal{S} \subsetneq [K] \mid \mathcal{S} \neq \emptyset\}}_{\text{\textsc{Partial}}}
\;\uplus\;
\underbrace{\{[K]\}}_{\text{\textsc{Full}}}.
\]

where $[K]=\{1,\dots,K\}$ denotes the index set of all segments, and $\mathcal{S}$ denotes a subset of segment indices, the selected blocks for aggregation.

These categories correspond to the three cases above and jointly encode both compression timing and intensity. \textsc{Null} ($u_t=\emptyset$) skips compression when the budget is sufficient. \textsc{Partial} selects a non-empty proper subset $\mathcal{S}$, where $|\mathcal{S}|$ controls the compression strength. \textsc{Full} ($u_t=[K]$) aggregates all segments under severe budget constraints.

Thus, a single action $u_t$ determines whether and how much to compress. After aggregation, the buffer is updated and the observation appended under the budget constraint. Because $\pi_\theta$ generates both compression decisions $u_t$ and reasoning actions in a unified action space, the agent learns to coordinate context reduction with downstream task performance. The policy therefore learns not only to solve the task under a fixed memory rule, but also to adjust context usage under changing budget conditions.

\subsection{Budget-Aware GRPO Objective with Progressive Context Curricula}
\label{sec:grpo_training}
To optimize the Agent for context management across varying budgets, we employ reinforcement learning under a progressive context budget curriculum. We utilize Group Relative Policy Optimization (GRPO) for sample efficiency without a value critic.

Formally, we define \(J\) curriculum stages with monotonically decreasing budgets \(B_{\max}^{(1)} \geq \dots \geq B_{\max}^{(J)}\). For each query, we sample \(N\) rollouts. Each rollout \(i\) is a full multi-turn interaction with \(M_i\) turns, where the model performs context management and generates responses across turns. The final outcome of rollout \(i\), denoted by \(\smash{R_i}\), is defined as the F1 score between the predicted answer and the ground truth. We assign rewards based on whether the entire rollout satisfies the stage-specific budget constraint. In the \(j\)-th curriculum stage, let \(|\mathcal{C}_{i,t}|\) denote the size of the managed context at turn \(t\) of rollout \(i\). The budget-constrained reward \(\tilde{R}_i^{(j)}\) and the group-relative advantage \(\smash{A_i^{(j)}}\) are computed as:
\begin{equation}
\tilde{R}_i^{(j)} =
\begin{cases}
R_i, & \text{if } \forall t,\ |\mathcal{C}_{i,t}| \leq B_{\max}^{(j)} \\
0, & \text{otherwise}
\end{cases},
\quad
A_i^{(j)} =
\smash{
\frac{
  \tilde{R}_i^{(j)} - \mathrm{mean}\bigl(\{\tilde{R}_n^{(j)}\}_{n=1}^N\bigr)
}{
  \mathrm{std}\bigl(\{\tilde{R}_n^{(j)}\}_{n=1}^N\bigr) + \epsilon
}
}.
\end{equation}

Each rollout produces token sequences across multiple turns, denoted as
$\{o_{i,m,t}\}_{t=1}^{T_{i,m}}$, where $T_{i,m}$ is the number of tokens
generated at turn $m$ in rollout $i$. We broadcast the trajectory-level advantage $\smash{A_i^{(j)}}$ to all tokens by
assigning the same advantage to each token within the trajectory and optimize the policy at the token level:

\begin{equation}
\mathcal{L}_{\text{PG}}(\theta) =
\frac{1}{N} \sum_{i=1}^{N}
\frac{1}{\sum_{m=1}^{M_i} T_{i,m}} 
\sum_{m=1}^{M_i} \sum_{t=1}^{T_{i,m}}
\min \left(
\smash{r_{i,m,t} A_i^{(j)}},
\smash{\mathrm{clip}(r_{i,m,t}, 1-\epsilon, 1+\epsilon) A_i^{(j)}}
\right)
\end{equation}

\begin{equation}
\mathcal{L}(\theta) =
\mathcal{L}_{\text{PG}}(\theta)
- \beta \mathbb{D}_{\mathrm{KL}}(\pi_{\theta} \| \pi_{\text{ref}})
\end{equation}

where $\smash{r_{i,m,t}} =
\frac{\pi_{\theta}(o_{i,m,t} \mid s_{i,m,t})}
{\pi_{\text{old}}(o_{i,m,t} \mid s_{i,m,t})}$ is the token-level probability ratio, and $s_{i,m,t}$ denotes the token state conditioned on the managed context at turn $m$.

This design provides a clear learning signal: only trajectories that succeed and respect the budget are rewarded, while group-relative normalization stabilizes optimization. By broadcasting trajectory-level advantages to all tokens, the model learns to align local decisions with globally effective context management under progressively tighter constraints.

%% file: section/Experiment.tex
\section{Experiments}
\label{sec:experiments}
This section evaluates Budget-Aware Context Management on compositional
multi-objective QA and long-horizon web browsing benchmarks. We introduce the datasets and metrics (Sections~\ref{sec:datasets}–\ref{sec:metrics}), followed by the compared methods (Section~\ref{sec:baselines}). We then present the main results (Section~\ref{sec:main-results}), including performance under different context budgets and compression behavior, and conclude with ablation studies (Section~\ref{sec:ablation-study}).

\subsection{Datasets}
\label{sec:datasets}
Existing multi-hop benchmarks are limited in horizon and do not explicitly require context management. Following MEM1 \citep{zhou2025mem1learningsynergizememory}, we aggregate multiple independent questions into a single composite query to form multi-objective tasks, increasing interaction length and reasoning complexity and inducing long-horizon trajectories. Consistent with MEM1, we restrict training data construction to 2-objective compositions, and evaluate generalization to larger compositions at test time.

For evaluation, we adopt the Wikipedia-based QA benchmarks used in Search-R1 \citep{jin2025searchr1}, covering single-hop datasets (NQ~\citep{kwiatkowski2019natural}, TriviaQA~\citep{joshi2017triviaqa}, PopQA~\citep{mallen2023popqa}) and multi-hop datasets (HotpotQA~\citep{yang2018hotpotqa}, 2WikiMultiHopQA~\citep{ho2020constructing}, MuSiQue~\citep{trivedi2022musique}, Bamboogle~\citep{press2022measuring}). Following MEM1 \citep{zhou2025mem1learningsynergizememory}, we convert them into multi-objective evaluation sets. We also include BrowseComp-Plus\citep{chen2025browsecompplus}, a long-horizon benchmark with a verified corpus, to evaluate performance in extended multi-step reasoning settings.

\subsection{Evaluation Metrics}
\label{sec:metrics}
For the multi-objective QA benchmarks, following MEM1 \citep{zhou2025mem1learningsynergizememory}, we compute token-level F1 between predicted and reference answers for each objective and report the sum of all objectives. For an $N$-objective task, the total score ranges from 0 to $N$, measuring answer accuracy and the ability to solve multiple sub-tasks within a single trajectory.

For BrowseComp-Plus, we follow the official LLM-as-a-Judge protocol~\citep{chen2025browsecompplus}, where correctness is determined by comparing model responses with ground-truth answers. Following the benchmark setup, we use Qwen3-32B as the judge model. Under this protocol, disagreement between Qwen3-32B and GPT-4.1 is below 1\%, and LLM judgments in BrowseComp-Plus baselines have been human-verified as reliable. We report average accuracy over all samples. Full judge prompts and details are provided in Appendix~\ref{app:judge}.

\subsection{Compared Methods}
\label{sec:baselines}
To evaluate Budget-Aware Context Management, we compare against the following baselines categorized by their context maintenance strategies:

\begin{itemize}
    \item[(1)] \textbf{Methods without Context Management}: We include ReAct \citep{yao2023reactsynergizingreasoningacting}, a reasoning-and-acting baseline without learned context control, and Search-R1 \citep{jin2025searchr1}, an RL-based search method without explicit context constraints.
    \item[(2)] \textbf{Reactive Context Management}: We employ the Summary Agent \citep{wu2025resumunlockinglonghorizonsearch, yu2025memagentreshapinglongcontextllm} as a conservative baseline that invokes summarization only when the context buffer is full. 
    \item[(3)] \textbf{Proactive Context Management}: We compare against MEM1 \citep{zhou2025mem1learningsynergizememory}, a reinforcement learning approach that maintains a compact, fixed-size context by iteratively consolidating past information into an internal state at each step.
\end{itemize}
Our method employs a curriculum learning strategy with progressive context budgets to stabilize learning under limited budgets, gradually tightening the context budget from 8k to 4k tokens during training to encourage adaptation as pressure increases. Detailed curriculum schedules are provided in Appendix~\ref{app:curriculum_details}.

For fair comparison, all methods are evaluated on two backbone models: Qwen2.5-7B-Instruct and Qwen3-30B-A3B-Instruct. We follow Search-R1's official implementation and training data (Appendix~\ref{app:search-r1}), and train the model under an 8k context budget for fair comparison. For the Summary Agent, we adopt a reactive summarization paradigm similar to ReSum and MemAgent, triggering summarization when the context window reaches a predefined threshold, the number of compressions is capped at 10 per task. For MEM1, we use the released 7B checkpoint and evaluation framework, adapting compression tags for the 30B backbone (Appendix~\ref{app:mem1}). Our method uses MEM1's multi-objective training data and limits compression to 10 operations per task. Additional details are provided in Appendix~\ref{app:hyperparams}. We plan to release our training code in future work to support reproducibility and facilitate follow-up research.

\subsection{Main Results}
\label{sec:main-results}

\paragraph{Performance Comparison.}
As shown in Table~\ref{tab:main-results}, BACM-RL achieves the best average performance on the BrowseComp-Plus and Multi-objective QA benchmarks. This holds across two backbone models with major architectural and parameter differences (7B and 30B), showing broad applicability. Compared to MEM1, the strongest prior baseline trained on the same data, BACM-RL achieves especially large gains in harder settings, including an approximate $5.0\times$ improvement in the 32-objective regime (4.545 vs. 0.909). This robustness extends to the out-of-domain BrowseComp-Plus benchmark, where our 30B variant, limited to an 8k budget, achieves 0.147 accuracy and surpasses the 235B Qwen3-Inst model (0.136) with a 128k context window. In the more context-demanding 32-objective regime, the large gap over its non-RL ablation (4.545 vs. 0.208) supports the effectiveness of our budget-aware GRPO objective design. By adopting a curriculum that progressively tightens the context budget, our framework guides the model to operate under increasingly strict constraints, encouraging local compression decisions that improve overall performance.

\begin{table}[t]
\small
\centering
\setlength{\tabcolsep}{3.8pt}
\renewcommand{\arraystretch}{1.15}
\resizebox{\linewidth}{!}{
\begin{tabular}{l c cccc cccc}
\toprule

\multirow{2}{*}{\textbf{Method}} &
\multirow{2}{*}{\textbf{Context}} &
\multicolumn{4}{c}{\textbf{BrowseComp-Plus}} &
\multicolumn{4}{c}{\textbf{Multi-objective QA}} \\

& &
Easy & Mid & Hard & Avg &
2-Obj & 8-Obj & 16-Obj & 32-Obj  \\

\midrule

\multicolumn{10}{l}{\textbf{Qwen3-235B-A22B-Inst-2507}} \\

ReACT & 128k & 0.380 & 0.164   & 0.034  & 0.136  & 0.948 & 1.873 & 1.315 & 1.412 \\
ReACT & 8k  & 0.420  & 0.143  & 0.011  & 0.118  & 0.886 & 1.782 & 1.233 & 0.374 \\

\midrule

\multicolumn{10}{l}{\textbf{Qwen2.5-7B-Inst}} \\

ReACT     & 8k & 0.320  & 0.098  & \uline{0.027}  & 0.089  & 0.395 & 0.458 & 0.376 & 0.090  \\
Search-R1(RL)$\ddagger$ & 8k & \uline{0.400}  & \uline{0.111}  & 0.015  & \uline{0.099}  & 0.760 & 1.719 & \uline{2.497} & 1.022 \\
Summary$^{\S}$   & 8k & 0.340  & 0.085  & 0.015  & 0.078  & 0.678 & 2.176 & 2.379 & 0.567  \\
MEM1(RL)$^{\dagger}$      & 8k & 0.040 & 0.044 & 0.015 & 0.035 & \uline{0.838} & \uline{2.345} & 2.391 & \uline{1.210}  \\
BACM (w/o RL) & 8k & 0.160  & 0.027  & 0.015  & 0.031  & 0.690 & 1.698 & 1.179 & 0.305  \\

\rowcolor{gray!15}
\textbf{BACM-RL} & 8k & \textbf{0.420}  & \textbf{0.146}  & \textbf{0.031}  & \textbf{0.127}  & \textbf{0.909} & \textbf{2.790} & \textbf{4.011} & \textbf{2.938}  \\

\midrule

\multicolumn{10}{l}{\textbf{Qwen3-30B-A3B-Inst}} \\

ReACT     & 8k & 0.440  & 0.148  & 0.034  & 0.130  & 0.938 & 2.078 & 0.931 & 0.208  \\
Search-R1(RL)$\ddagger$ & 8k & 0.420  & 0.150  & 0.027  & 0.128  & \uline{1.013} & \uline{3.310} & 1.949 & 0.998 \\
Summary$^{\S}$   & 8k & \uline{0.480}  & \textbf{0.164}  & 0.019  & \uline{0.137}  & 0.916 & 2.456 & \uline{2.992} & \uline{2.848}  \\
MEM1$\ddagger$      & 8k & 0.360 & 0.141 & \textbf{0.069} & 0.131 & 0.978 & 1.327 & 1.383 & 0.909  \\
BACM (w/o RL) & 8k & 0.200  & 0.100  & 0.015  & 0.080  & 0.474 & 0.098 & 0.170 & 0.208  \\
\rowcolor{gray!15}
\textbf{BACM-RL} & 8k & \textbf{0.520}  & \textbf{0.164}  & \uline{0.042}  & \textbf{0.147}  & \textbf{1.032} & \textbf{3.587} & \textbf{6.255} & \textbf{4.545}  \\

\bottomrule
\end{tabular}
}
\footnotesize
\caption{
Performance comparison across QA settings.
BrowseComp-Plus is evaluated using \textit{LLM-as-Judge}, while Multi-object QA reports \textit{F1 scores summed across objectives}.$\dagger$ Reproduced using officially released open-source checkpoints.;
$\ddagger$ Re-implemented based on released code;
$\S$ Re-implemented based on the original paper due to no public implementation.}
\label{tab:main-results}
\end{table}

\begin{figure}[t] 
    \centering
    \vspace{-5pt}
    \includegraphics[width=1.0\linewidth]{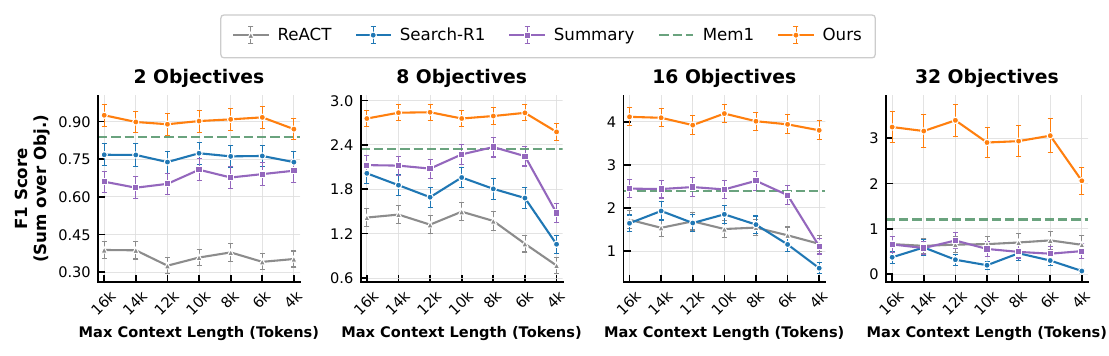}
    \vspace{-20pt}
    \caption{Performance under different maximum context window sizes (16k-4k tokens) with varying numbers of objectives.}
    \label{fig:f1_vs_context}
    \vspace{-0pt}
\end{figure}

\vspace*{-7pt} 
\paragraph{Robustness to Varying Context Budgets}
Figure~\ref{fig:f1_vs_context} demonstrates the method's superiority and stability under all task complexities and extensive context budget settings. Unlike context-management-free baselines (ReACT, RL-based Search-R1), which compete only on simple tasks with sufficient budgets, our approach remains nearly invariant under budget reductions from 16k to 4k tokens. While budget-free strategies like MEM1 and Summary partially alleviate degradation, they remain inferior. MEM1 incurs information loss via turn-by-turn over-compression, while Summary suffers reasoning failures by delaying compression. By contrast, formulating context management as a budget-aware \textit{sequential decision problem} enables adaptive modulation of commit-block aggregation intensity. This is most pronounced in the extreme 32-objective setting at the 4k limit, where the method achieves a 1.7$\times$ cumulative F1 improvement over the best baseline (2.06 vs. 1.21).


\paragraph{Compression Efficiency Under Budget Constraints}

\begin{wrapfigure}{r}{0.40\textwidth}
\centering
\includegraphics[width=1.0\linewidth]{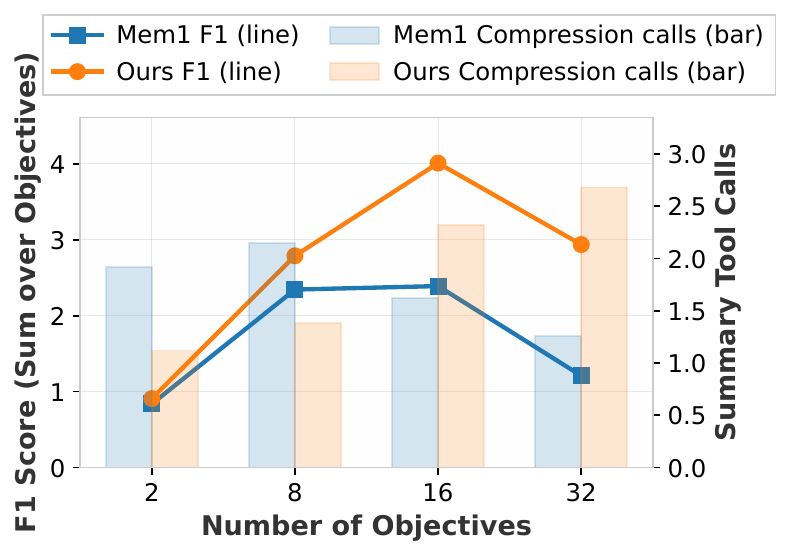}
\vspace{-20pt}
\caption{Cumulative F1 and average compression calls under a fixed 8k context budget.}
\label{fig:Summary_calls}
\vspace{-10pt}
\end{wrapfigure}


Figure~\ref{fig:Summary_calls} indicates that under a fixed 8k budget, our gains reflect budget-aware management rather than indiscriminately increasing compression frequency. Under light workloads (2/8 objectives), our method improves F1 over MEM1 by 8.3\%/18.7\% (0.84→0.91, 2.35→2.79) while reducing compression calls by 41.7\%/35.8\% (1.92→1.12, 2.15→1.38). At 16/32 objectives, compression calls rise by 43\%/109\% (1.62→2.32, 1.28→2.68), with F1 improving by 67\%/143\% (2.40→4.01, 1.21→2.94). MEM1 fails because of context saturation and signal loss, causing premature responses. Meanwhile, our method balances token usage and performance; token statistics are in Appendix~\ref{app:efficiency_tradeoff}.


\subsection{Ablation Study}
\label{sec:ablation-study}

\paragraph{Impact of Budget-Aware State}
\begin{wraptable}{r}{0.38\textwidth}
\centering
\small
\vspace{-15pt}
\begin{tabular}{@{}lcc@{}}
\toprule
\textbf{Variant} & \textbf{Comp.} & \textbf{Budget} \\ \midrule
Search-R1 (Base)             & $-$            & $-$             \\
Search-R1 (w/ B) & $-$            & $\checkmark$    \\
Ours (w/o B)     & $\checkmark$   & $-$             \\
\rowcolor{gray!15}
\textbf{Ours (Full)} & $\checkmark$ & $\checkmark$  \\ \bottomrule
\end{tabular}
\vspace{-5pt}
\caption{Ablation configurations.}
\label{tab:ablation}
\vspace{-5pt}
\end{wraptable}
To analyze the framework's performance contributions, we evaluate four configurations by decoupling budget metadata from the compression policy (Table~\ref{tab:ablation}). We define \textit{Base} (Search-R1) as the vanilla model without context management. \textit{Search-R1 (w/ B)} adds budget metadata to the agent state, while \textit{Ours (w/o B)} uses the learned compression policy without explicit budget signals. \textit{Ours (Full)} integrates both for budget-conditioned context optimization.

\begin{figure}[t]
\centering
\includegraphics[width=1\linewidth]{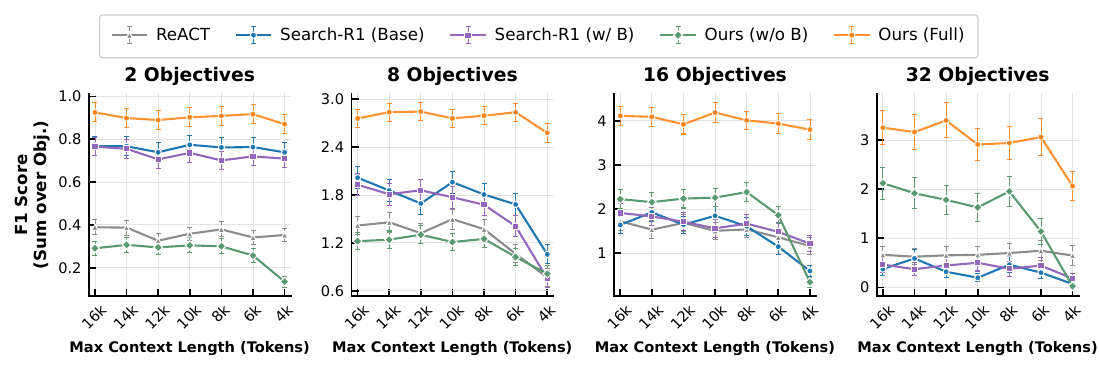}
\vspace{-20pt}
\caption{ Ablation of key components across context budgets (16k–4k tokens), measured by summed F1 across objectives. Removing budget metadata (B) degrades performance (\textit{Ours w/o B}), while adding B alone to a baseline without context management does not improve performance (\textit{Search-R1 w/ B}). Our full model (\textit{Ours Full}), which conditions compression on B, achieves the best performance across all objectives and budgets.
}
\label{fig:context_budget_results}
\end{figure}

Figure~\ref{fig:context_budget_results} shows that removing budget information from our compression mechanism causes significant performance degradation across context budgets and multi-objective settings. While \textit{Search-R1 (w/ B)} yields only marginal gains, our full framework consistently achieves the highest cumulative F1 scores. By explicitly integrating budget signals, our approach enables superior dynamic trade-offs between preserving raw context fidelity and invoking aggressive summarization as the budget diminishes.

\begin{wraptable}{r}{0.38\textwidth}
\centering
\small
\vspace{-10pt}
\setlength{\tabcolsep}{1.2pt}
\renewcommand{\arraystretch}{1.02}
\begin{tabular}{@{}l@{\hspace{3pt}}cccc@{}}
\toprule
\textbf{Variant} & \textbf{2-Obj} & \textbf{8-Obj} & \textbf{16-Obj} & \textbf{32-Obj} \\ \midrule
w/o Def. & 0.833 & 2.706 & 3.972 & 2.818 \\
\rowcolor{gray!15}
w/ Def. & \textbf{0.909} & \textbf{2.790} & \textbf{4.011} & \textbf{2.938} \\
\bottomrule
\end{tabular}
\caption{Cumulative F1 of deferred vs.\ non-deferred loading across different objectives at $8\text{k}$}
\label{tab:ablation_deferred_loading}
\vspace{-15pt}
\end{wraptable}


To further analyze our deferred-loading mechanism, we compare it with the pre-loading strategy under the same budget-constrained multi-objective setting (Table~\ref{tab:ablation_deferred_loading}). The deferred version accounts for the upcoming observation size in the budget-aware state before loading. In contrast, pre-loading incorporates the observation without such awareness. Deferred loading consistently outperforms pre-loading, demonstrating the benefit of budget-aware deferred loading.

\paragraph{Ablation of Progressive Context Budget Curriculum}


To isolate the effect of the progressive context budget curriculum, we ablate on Qwen2.5-7B-Instruct with three training schedules: a static $8\text{k}$ context budget, a random context budget sampled from $\{4\text{k},8\text{k}\}$, and a progressive context budget schedule of $8\text{k}\!\rightarrow\!4\text{k}$, all evaluated under the same maximum context budget of $8\text{k}$ (see Appendix~\ref{app:training_dynamics} for more training dynamics).

\begin{wrapfigure}{r}{0.44\textwidth}
\centering
\vspace{-15pt}
\includegraphics[width=1\linewidth]{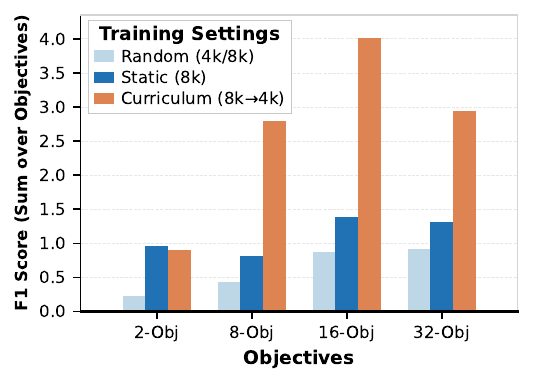}
\vspace{-25pt}
\caption{Ablation of progressive budget curricula under a common 8k evaluation budget summed F1 over objectives}
\vspace{-15pt}
\label{fig:ablation_curriculum}
\end{wrapfigure}

Figure~\ref{fig:ablation_curriculum} shows the progressive curriculum is critical for budget-aware context management. Randomized training performs worst, indicating unstructured budget variation hinders learning. Although static $8\text{k}$ training performs well in the 2-objective setting ($0.967$), it scales poorly, reaching $1.391$ and $1.319$ on 16 and 32 objectives. In contrast, the progressive curriculum achieves the best overall performance ($2.662$ average), with strong gains on complex tasks, scoring $4.011$ and $2.938$ on 16 and 32 objectives. These results show progressively increasing budget pressure is essential for generalizing context-management behaviors to long-horizon settings.

\subsection{Case Study: Analysis of Model Behavior}
We analyze behavior under \texttt{[NONE]}, \texttt{[Selective]}, and \texttt{[ALL]} (see App.~\ref{fig:case_study}). With sufficient budget (e.g., $\sim$45.7\%), the agent adopts \texttt{NONE}, preserving \texttt{commit\_blocks} and deferring compression. Under moderate pressure (e.g., $\sim$30.6\%), it switches to \texttt{Selective}, merging earlier blocks (e.g., \texttt{commit\_ids: "c1,c2,c3"}) to condense history while retaining newer, task-relevant details (e.g., \texttt{c4} reorganized as \texttt{c2}, \texttt{c3}). When the budget is tight (e.g., $\sim$28.7\%), it escalates to \texttt{ALL}, invoking \texttt{commit\_ids: "ALL"} to compress history into a \texttt{merged\_summary}, increasing compression to meet the budget before appending new observations.

%% file: section/Conclusions.tex
\vspace{-5pt}
\section{Conclusion}
\vspace{-5pt}
We study context management for long-horizon LLM agents under strict context constraints and show that budget-agnostic compression leads to information loss or overflow. We propose BACM, which formulates compression as a budget-conditioned sequential decision problem and optimizes it via reinforcement learning (BACM-RL). Experiments demonstrate consistent improvements in robustness across context budgets, especially under tight constraints. Overall, our results highlight the importance of budget-aware context management for reliable long-horizon reasoning.

%% file: section/Appendix.tex
\section{Appendix}

\subsection{Limitations and future work}

Despite the strong empirical performance achieved by formulating context management as a budget-aware sequential decision problem, several limitations remain that warrant further investigation. (1) The reliance on trajectory-level reinforcement learning introduces challenges associated with sparse and delayed reward signals, which may limit the effectiveness of process-level supervision for context management decisions; a promising direction is to incorporate more structured credit assignment mechanisms, such as intermediate supervision, hierarchical reward design, or explicit modeling of long-term information utility \citep{feng2025gigpo}. (2) While commit-block aggregation enables flexible and adaptive control over compression under varying budget conditions, it operates at a coarse segment level; exploring finer-grained importance modeling, such as token- or span-level saliency estimation, may further improve the preservation of critical information in tasks requiring high factual precision. (3) Although the evaluation demonstrates clear improvements on multi-objective QA and long-horizon browsing benchmarks designed to stress-test context management, extending the framework to more realistic and diverse environments, including open-ended tool use, multimodal reasoning, and human-agent interaction, would provide a more comprehensive assessment of generalization and practical applicability.

Overall, budget-aware context management provides a promising direction for enabling robust long-horizon reasoning under strict resource constraints, and future advances along these directions may further enhance its effectiveness in real-world agent systems.

\input{section/case/case_study_latex}

\subsection{Efficiency--Performance Trade-offs with Improved Context Utilization}
\label{app:efficiency_tradeoff}

\begin{table}[H]
\footnotesize
\setlength{\tabcolsep}{2.5pt}
\renewcommand{\arraystretch}{1.05}
\resizebox{\columnwidth}{!}{%
\begin{tabular}{l c | ccc | ccc | ccc | ccc}
\toprule
\multirow{2}{*}{Method} & \multirow{2}{*}{Ctx}
& \multicolumn{3}{c|}{2 Obj}
& \multicolumn{3}{c|}{8 Obj}
& \multicolumn{3}{c|}{16 Obj}
& \multicolumn{3}{c}{32 Obj} \\
\cmidrule(lr){3-5} \cmidrule(lr){6-8} \cmidrule(lr){9-11} \cmidrule(lr){12-14}
& & D & P & S & D & P & S & D & P & S & D & P & S \\
\midrule
Summary & 16k
& 664.469 & 1517.570 & 0.661
& 1451.613 & 3871.188 & 2.127
& 1658.180 & 4938.746 & 2.447
& 2894.227 & 6306.773 & 0.656 \\

Search-R1 & 16k
& \underline{306.648} & \underline{792.320} & 0.768
& 1481.602 & 3769.055 & 2.016
& 2280.371 & 5729.789 & 1.643
& 3669.387 & 6810.332 & 0.372 \\

Mem1 & 16k
& \textbf{304.439} & \textbf{632.912} & 0.838
& \textbf{652.492} & \textbf{914.818} & 2.345
& \textbf{876.083} & \textbf{1096.275} & 2.391
& \textbf{1046.315} & \textbf{1361.961} & 1.210 \\

Ours & 16k
& 445.133 & 2032.777 & \textbf{0.925}
& \underline{1227.461} & \underline{2986.828} & \textbf{2.756}
& \underline{1919.848} & \underline{3537.504} & \textbf{4.114}
& 7600.406 & 8795.160 & \textbf{3.249} \\

\midrule
Search-R1 & 8k
& \underline{304.152} & \underline{793.641} & 0.762
& 1332.250 & 3445.043 & 1.805
& 1920.008 & 4952.941 & 1.612
& 2556.023 & 5065.641 & 0.461 \\

Summary & 8k
& 409.352 & 1234.168 & 0.677
& 758.180 & 2780.047 & \underline{2.371}
& 1027.000 & 4160.898 & \underline{2.628}
& 1508.281 & 4497.906 & 0.491 \\

Mem1 & 8k
& \textbf{304.439} & \textbf{632.912} & 0.838
& \textbf{652.492} & \textbf{914.818} & 2.345
& \textbf{876.083} & \textbf{1096.275} & 2.391
& \textbf{1046.315} & \textbf{1361.961} & 1.210 \\

Ours & 8k
& 482.633 & 2069.656 & \textbf{0.909}
& \underline{1030.082} & \underline{2767.555} & \textbf{2.790}
& \underline{1848.777} & \underline{3442.949} & \textbf{4.011}
& 5224.945 & 6436.852 & \textbf{2.938} \\

\midrule
Search-R1 & 4k
& \underline{298.672} & \underline{763.586} & 0.739
& 910.277 & 2849.121 & 1.056
& 1145.297 & 3522.727 & 0.600
& 1486.727 & 3317.695 & 0.069 \\

Summary & 4k
& 508.785 & 2710.801 & \underline{1.479}
& 858.246 & 3087.055 & 0.508
& 305.969 & 1192.008 & 0.704
& 701.258 & 3448.832 & 1.109 \\

Mem1 & 4k
& \textbf{304.439} & \textbf{632.912} & 0.838
& \textbf{652.492} & \textbf{914.818} & 2.345
& \textbf{876.083} & \textbf{1096.275} & 2.391
& \textbf{1046.315} & \textbf{1361.961} & 1.210 \\

Ours & 4k
& 403.254 & 1985.531 & \textbf{0.870}
& \underline{942.934} & \underline{2659.348} & \textbf{2.576}
& \underline{1584.344} & \underline{3129.082} & \textbf{3.801}
& \underline{2862.035} & \underline{4001.288} & \textbf{2.061} \\

\bottomrule
\end{tabular}%
}
\caption{mean dependent cost (D), mean peak tokens (P), and summed F1 (S) across context budgets and objective counts.}
\label{tab:cost_peak_score_comparison}
\vspace{-4pt}
\end{table}

We compare methods under multiple context budgets (4k, 8k, 16k) and varying objective counts, reporting mean dependent cost (D), peak context tokens (P), and summed F1 (S). As shown in Table~\ref{tab:cost_peak_score_comparison}, the proposed method adapts token usage to both task complexity and available context, and consistently achieves the highest performance.

For simple tasks (2 objectives), token usage remains stable across context budgets (e.g., Peak: 2033 at 16k vs.\ 1986 at 4k) while yielding higher scores, indicating that additional context does not introduce unnecessary token growth. As task complexity increases (8--16 objectives), the proposed method maintains superior performance with competitive or lower Peak tokens compared to methods without effective context control (e.g., 16 objectives: 3538 vs.\ 4939 for Summary at 16k; 3443 vs.\ 4161 at 8k), which reflects more efficient utilization of context. For more complex settings (32 objectives), Peak tokens increase with larger context budgets (4339 $\rightarrow$ 6437 $\rightarrow$ 8795 from 4k to 16k), leading to consistent performance improvements. In contrast, Mem1 maintains minimal token usage (1362) but exhibits limited performance gains, primarily because it does not preserve sufficient task-relevant information within the available context.

Overall, these results indicate that effective context management requires adaptive allocation of tokens rather than aggressive compression or delayed summarization.

\subsection{Answer rate analysis across context lengths}
\label{app:answer_rate}

\begin{figure*}[t]
    \centering
    \includegraphics[width=\textwidth]{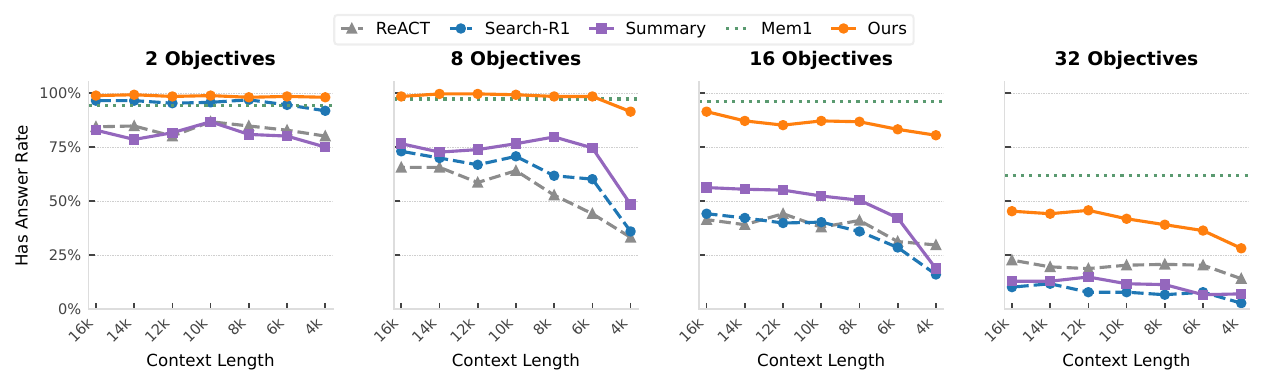}
    \caption{Answer rate comparison across context lengths and objectives.}
    \label{fig:has_answer_rate}
\end{figure*}

Figure~\ref{fig:has_answer_rate} shows answer rates across context lengths (4K--16K tokens) and task complexities (2/8/16/32 objectives) on the 7B model.

In simple scenarios (2--8 objectives), all methods exhibit answer rates above 90\%. At higher complexities (16--32 objectives), baseline methods (ReACT, Search-R1, Summary) show lower answer rates, particularly at shorter context lengths. Mem1 maintains higher answer rates across settings (61.7\% at 32 objectives).

In scenarios with 2--16 objectives, our method exhibits answer rates comparable to Mem1. At 32 objectives, our method shows a lower answer rate than Mem1, while corresponding to higher task performance due to retaining more information; compared to other baselines, our method maintains higher answer rates. Task performance results are reported in the main experiments.

\subsection{Implementation Details and Hyperparameters} \label{app:hyperparams}
\subsubsection{Training infrastructure and Core Hyperparameters}
All experiments are conducted using the \texttt{verl} framework. Our method and the Search-R1 baseline share a unified RL configuration for consistency. We employ Group Relative Policy Optimization (GRPO) with a total of 16 nodes, each equipped with 8 GPUs. The training is performed for a total of 300 steps with a global batch size of 128. For the actor model, we use a cosine learning rate schedule starting at with 100 warmup steps. Detailed hyperparameters are summarized in Table~\ref{tab:hyperparams}.

\subsubsection{Curriculum Reinforcement Learning Implementation}
\label{app:curriculum_details}
To enhance the model's context-management capabilities under progressively constrained resources, our method adopts a curriculum-based reinforcement learning strategy. Training is partitioned into \(K=5\) stages, each consisting of exactly 60 steps (yielding a total of 300 steps).

The maximum model length (\texttt{max\_model\_len}) follows the curriculum schedule detailed in Table~\ref{tab:curriculum}. It starts at 8,192 tokens in Stage 1 and is linearly reduced to 4,096 tokens in Stage 5 (a uniform decrease of 1,024 tokens per stage).

Unlike prior approaches that utilize a fixed maximum context length, our curriculum gradually tightens the context-window budget \(\smash{B_{\max}^{(k)}}\) across \(\smash{K}\) training stages, where \(\smash{k \in \{1, \dots, K\}}\) denotes the current stage. This mechanism forces the agent to internalize increasingly sophisticated management behaviors as it transitions from ample capacity to extreme scarcity.

To ensure fair comparison, the total training steps for our method as well as all baselines are uniformly fixed at 300 steps.

\begin{table}[t]
\centering
\label{tab:curriculum}
\begin{tabular}{lccccc}
\toprule
 & \textbf{Stage 1} & \textbf{Stage 2} & \textbf{Stage 3} & \textbf{Stage 4} & \textbf{Stage 5} \\
\midrule
\textbf{Step Range}          & 1--60   & 61--120 & 121--180 & 181--240 & 241--300 \\
\textbf{Max Model Length (tokens)} & 8,192 & 7,168 & 6,144 & 5,120 & 4,096 \\
\bottomrule
\end{tabular}
\caption{Curriculum Schedule for \texttt{max\_model\_len}}
\end{table}


\subsubsection{Search-R1 Implementation} \label{app:search-r1}
For the Search-R1 baseline, we strictly follow the official implementation and training data construction as described in \citet{jin2025searchr1}. The model is trained to utilize external search tools within the reasoning chain. The environment interface and multi-turn interaction format (Hermes) are kept identical to our proposed method to ensure a fair comparison of context management capabilities. All training is performed for exactly 300 steps under the shared RL configuration (with fixed \texttt{max\_model\_len}=8,192), by which point training has largely converged.

\subsubsection{MEM1 Implementation} \label{app:mem1}
The MEM1 baselines utilize the official implementation provided by \citet{zhou2025mem1learningsynergizememory}. For the Qwen2.5-7B-Instruct model, we use the provided weights and prompt templates. 
We found that the 30B-A3B model responded more effectively to modified special tokens, thus, we replaced the standard \texttt{<think>} tags with \texttt{<internal\_state>} tags in the prompt to ensure the model correctly triggered its reasoning and memory behaviors. 

\begin{table}[t]
\centering
\setlength{\tabcolsep}{4pt}
\renewcommand{\arraystretch}{0.95}
\begin{tabular}{>{\centering\arraybackslash}p{3.5cm} >{\centering\arraybackslash}p{5cm}}
\toprule
\textbf{Hyperparameter} & \textbf{Value} \\
\midrule
Optimizer & AdamW \\
Actor Learning Rate & $1\times10^{-6}$ \\
LR Schedule & Cosine \\
Warmup Steps & 100 \\
Rollout Group Size ($G$) & 5 \\
KL Coefficient ($\beta$) & 0.001 \\
GRPO Mini-batch Size & 128 \\
Max Model Length & 8,192 (initial; curriculum reduces to 4,096 for our method) \\
\bottomrule
\end{tabular}
\caption{Global Hyperparameters for RL Training}
\label{tab:hyperparams}
\end{table}

\subsection{Budget-Aware Context Management Prompt}
\label{app:budget_prompt}
The prompt variables are computed from the current agent state before the pending tool response is appended. Here, \texttt{current\_ctx\_len} is the token length of the current prompt buffer, and \texttt{tool\_response\_len} is the token length of the deferred tool response text. We conservatively set \texttt{usable\_limit} to \texttt{max\_model\_len} minus a 1{,}000-token safety margin. The remaining budget is then \texttt{usable\_limit} minus the projected post-load length, i.e., \texttt{current\_ctx\_len} + \texttt{tool\_response\_len}, and \texttt{remaining\_pct} is its normalized percentage. This budget prompt is used only when a later tool response is deferred, so folding decisions are made before the new observation is loaded.

\begin{center}
\begin{minipage}{0.98\linewidth}
\begin{tcolorbox}[
  colback=gray!5!white,
  colframe=gray!75!black,
  title=Budget State Prompt,
  boxrule=0.5pt,
  arc=3pt,
  left=2mm,
  right=2mm,
  top=1mm,
  bottom=1mm
]
\small
\texttt{[Remaining context capacity prior to receiving tool call results]} \\
Please decide whether to summarize or fold the upcoming tool response to stay within limits. \\
- Current prompt length: \{current\_ctx\_len\} tokens \\
- Estimated tool response length: \{tool\_response\_len\} tokens \\
- Remaining tokens for next turn: \{remaining\_budget\} tokens (\{remaining\_pct:.1f\}\% of usable context) \\
- Usable context limit (max length minus 1{,}000 safety margin): \{usable\_limit\} tokens \\
Decide whether to fold previous commits to save context budget. \\
OPTIONS: \\
- ``NONE'': Keep all commits (default) \\
- ``ALL'': Fold everything (only if context is low) \\
- ``c0001,c0002'': Fold specific old commits \\
RULE: Don't fold unless necessary. Preserve user requirements and errors. \\
OUTPUT format: \\
\texttt{<tool\_call>} \\
\texttt{\{"name": "summarize", "arguments": \{"fold\_commit\_ids": "NONE", "merged\_commit": ""\}\}} \\
\texttt{</tool\_call>} \\
Examples: \\
- Safe: \texttt{fold\_commit\_ids="NONE", merged\_commit=""} \\
- Low context: \texttt{fold\_commit\_ids="ALL", merged\_commit="[key points from session]"} \\
- Selective: \texttt{fold\_commit\_ids="c0001,c0002", merged\_commit="[merged content]"}
\end{tcolorbox}
\label{fig:budget_state_prompt}
\end{minipage}
\end{center}
Ours (w/o B) uses the learned compression policy without explicit budget signals. Compared to the Budget State Prompt, it removes all budget-related information.
\begin{center}
\begin{minipage}{0.98\linewidth}
\begin{tcolorbox}[
  colback=gray!5!white,
  colframe=gray!75!black,
  title=Compression Prompt (w/o Budget),
  boxrule=0.5pt,
  arc=3pt,
  left=2mm,
  right=2mm,
  top=1mm,
  bottom=1mm
]
\small
\texttt{"Please choose the most appropriate folding strategy"} \\
\texttt{"The options include: Partial (e.g., c0001,c0002) / ALL / NONE."} \\
\texttt{"'c0001,c0002' means folding those commits and replacing them with merged\_commit information."} \\
\texttt{"'ALL' means folding all commits and replacing them with merged\_commit."} \\
\texttt{"'NONE': Keep all commits (default)"} \\
OUTPUT format: \\
\texttt{<tool\_call>} \\
\texttt{\{"name": "summarize", "arguments": \{"fold\_commit\_ids": "NONE", "merged\_commit": ""\}\}} \\
\texttt{</tool\_call>}
\end{tcolorbox}
\label{fig:wo_budget_prompt}
\end{minipage}
\end{center}

\subsection{LLM-as-a-Judge Evaluation Details}
\label{app:judge}
We follow the BrowseComp-Plus evaluation protocol and implement the judge using the Qwen3-32B model. The judge receives the question, the model-generated response, and the ground-truth answer, and determines whether the predicted answer is semantically equivalent to the reference answer.

Following the benchmark design, two independent judges with identical prompts are used to assess answer correctness, and their agreement differs by less than 1\%, indicating high evaluation reliability. The BrowseComp-Plus authors further report that LLM-judged results closely match human annotations.

The judge prompt used in our evaluation is shown in Figure~\ref{fig:judge_prompt}.

\begin{tcolorbox}[
  colback=gray!5!white,
  colframe=gray!75!black,
  title=Judge Prompt,
  boxrule=0.5pt,
  arc=3pt,
  left=2mm,
  right=2mm,
  top=1mm,
  bottom=1mm
]
\small
Judge whether the following response to a question is correct based on the provided correct answer.

\textbf{Question:} \{question\} \\
\textbf{Response:} \{response\} \\
\textbf{Correct Answer:} \{correct\_answer\} \\
\textbf{Instructions:}

1. Extract the final answer from the response. \\
2. Determine whether the extracted answer is semantically equivalent to the correct answer. Minor wording differences are acceptable. The extracted answer may be more detailed than the correct answer as long as the additional information is correct. \\
3. Output the following fields: \\

\texttt{extracted\_final\_answer}: the final answer extracted from the response \\
\texttt{correct\_answer}: repeat the provided correct answer \\
\texttt{reasoning}: explanation of whether the extracted answer matches the correct answer \\
\texttt{correct}: yes or no \\
\texttt{confidence}: value between 0 and 100
\label{fig:judge_prompt}
\end{tcolorbox}

\subsection{Training Dynamics Analysis}
\label{app:training_dynamics}

\begin{figure}[H]
\centering
\includegraphics[width=\textwidth]{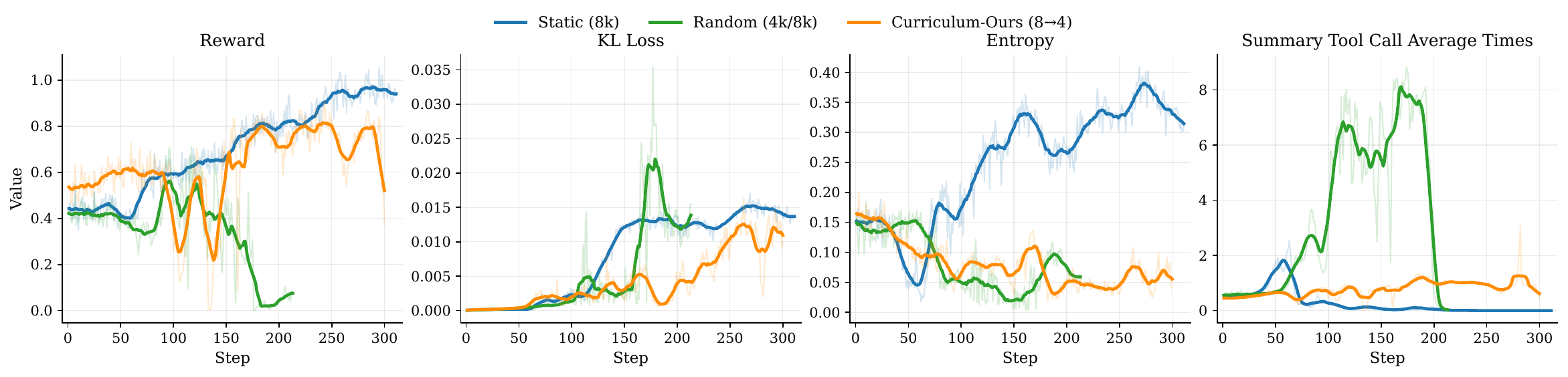}
\caption{Training dynamics of Qwen2.5-7B-Instruct across Static (8k), Random (4k/8k), and Curriculum-Ours (8$\rightarrow$4): Reward, KL Loss, Entropy, and Summary Tool Call Average Times.}
\label{fig:training_dynamics}
\end{figure}

\textbf{Training setup.} Figure~\ref{fig:training_dynamics} shows the training dynamics of Qwen2.5-7B-Instruct under three context-budget schedules. The Static (8k) baseline keeps the maximum context budget fixed at 8k throughout training, representing the original setting. Our Curriculum (8$\rightarrow$4) progressively reduces the maximum context budget from 8k to 4k as training proceeds, with the budget tightened every 60 steps. To control for the effect of merely exposing the model to a 4k budget, we further introduce a Random (4k/8k) schedule, in which the maximum context budget at each step is randomly sampled from $\{4\text{k}, 8\text{k}\}$.

The Static (8k) baseline achieves stable reward improvement throughout training, while the Random (4k/8k) strategy fails to converge, exhibiting reward collapse after step 180 accompanied by KL divergence spikes. In contrast, our progressive curriculum maintains consistent reward gains even during steps 200--300 under tightening constraints. The summary-tool call frequency reveals distinct compression behaviors: Static training shows near-zero compression after step 100, as abundant context removes compression pressure, whereas our curriculum progressively increases compression frequency as the budget decreases from 8k to 4k. This confirms that gradually tightening context budgets effectively induces adaptive context-management behaviors, enabling the model to learn compression strategies rather than avoiding them when resources are plentiful.

%% file: section/case/case_study_latex.tex
\subsection{Case study of budget-aware context management}
\label{app:case_study}
\begin{figure}[t] 
    \centering
    \vspace{-10pt}
    \includegraphics[width=1.0\linewidth]{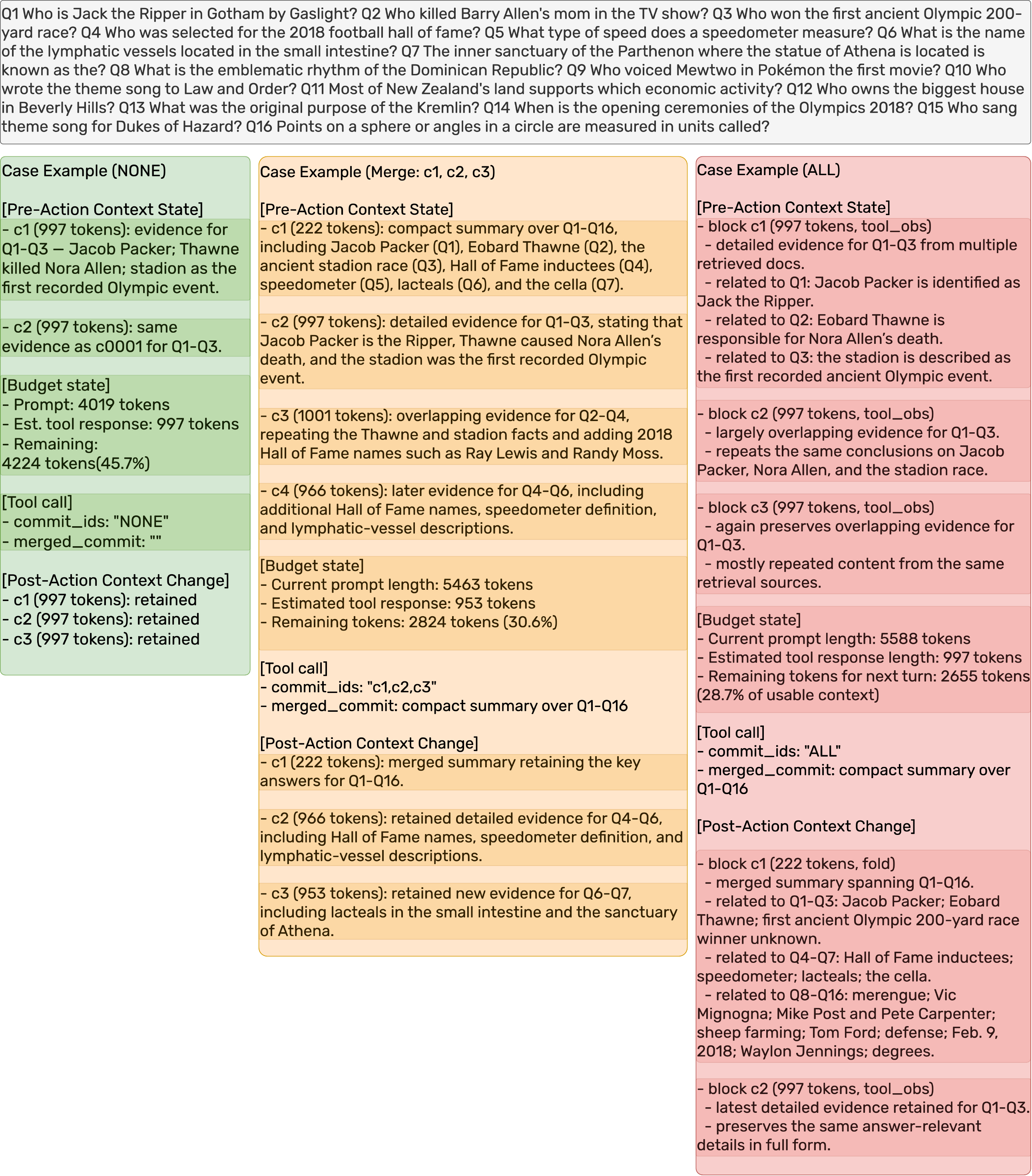}
    \vspace{-20pt}
    \caption{Case study of budget-aware context management. The agent adaptively selects \texttt{[NONE]}, \texttt{[Selective]}, or \texttt{[ALL]} based on remaining budget, transitioning from full retention to partial and full aggregation as budget tightens, while preserving task-relevant information.}
    \label{fig:case_study}
    \vspace{-0pt}
\end{figure}

We provide additional details for the case study presented in the main text (Figure~\ref{fig:case_study}). The example illustrates how the agent dynamically adjusts context management strategies under varying budget conditions, selecting among \texttt{[NONE]}, \texttt{[Selective]}, and \texttt{[ALL]}.



